\documentclass[letterpaper, 10 pt, journal, twoside]{IEEEtran}

\IEEEoverridecommandlockouts                              

\usepackage[top=60pt, left=48pt, right=48pt, bottom=43pt]{geometry}

\usepackage{graphicx} 

\usepackage{times}

\usepackage{multicol}
\usepackage[bookmarks=true]{hyperref}
\usepackage{amssymb}
\usepackage{color}
\definecolor{dgreen}{rgb}{0,0,0}
\definecolor{dyellow}{rgb}{.7,.7,0}
\definecolor{dred}{rgb}{1,0,0}
\definecolor{dblue}{rgb}{0,0,0.7}
\definecolor{dorange}{rgb}{0.9,0.5,0.1}

\usepackage[acronym]{glossaries}
\newacronym{coolname}{HULC}{Hierarchical Universal Language Conditioned Policies}
\usepackage{xcolor}
\usepackage{pifont}
\usepackage{colortbl}
\newcommand*\colourcheck[1]{%
  \expandafter\newcommand\csname #1check\endcsname{\textcolor{#1}{\ding{52}}}%
}
\newcommand*\colourcross[1]{%
  \expandafter\newcommand\csname #1xmark\endcsname{\textcolor{#1}{\ding{55}}}%
}
\colourcheck{green}
\colourcross{red}
\definecolor{Gray}{gray}{0.9}
\definecolor{LightCyan}{rgb}{0.88,1,1}
\usepackage{cite}
\begin{document}

\title{What Matters in Language Conditioned Robotic\\ Imitation Learning over Unstructured Data}

\author{Oier Mees$^{*1}$, Lukas Hermann$^{*1}$, Wolfram Burgard$^{2}$\\ 
 \thanks{Manuscript received: February, 24, 2022; Accepted July, 11, 2022.}
\thanks{This paper was recommended for publication by Editor Dana Kulic upon evaluation of the Associate Editor and Reviewers' comments.}
 \thanks{$^\ast$Equal contribution.$^{1}$University of Freiburg, Germany.$^{2}$University of Technology Nuremberg, Germany. {\tt\footnotesize meeso@informatik.uni-freiburg.de}}
\thanks{Digital Object Identifier (DOI): see top of this page.}
\thanks{© 2022 IEEE.  Personal use of this material is permitted.  Permission from IEEE must be obtained for all other uses, in any current or future media, including reprinting/republishing this material for advertising or promotional purposes, creating new collective works, for resale or redistribution to servers or lists, or reuse of any copyrighted component of this work in other works.}
}

\maketitle

\markboth{IEEE Robotics and Automation Letters. Preprint Version. Accepted July, 2022}
{Mees \MakeLowercase{\textit{et al.}}: What Matters in Language Conditioned Robotic Imitation Learning over Unstructured Data} 

\begin{abstract}
A long-standing goal in robotics is to build robots that can perform a wide range of daily tasks from perceptions obtained with their onboard sensors and specified only via natural language. While recently substantial advances have been achieved in language-driven robotics by leveraging end-to-end learning from pixels, there is no clear and well-understood process for making various design choices due to the underlying variation in setups.
In this paper, we conduct an extensive study of the most critical challenges in learning language conditioned policies from offline free-form imitation datasets.
We further identify architectural and algorithmic techniques that improve performance, such as a hierarchical decomposition of the robot control learning, a multimodal transformer encoder, discrete latent plans and a self-supervised contrastive loss that aligns video and language representations.
By combining the results of our investigation with our improved model components, we are able to present a novel approach that significantly outperforms the state of the art on the challenging language conditioned long-horizon robot manipulation CALVIN benchmark.
We have open-sourced our implementation to facilitate future research in learning to perform many complex manipulation skills in a row specified with natural language. Codebase and trained models available at \url{http://hulc.cs.uni-freiburg.de}
\end{abstract}
\begin{IEEEkeywords}
Learning Categories and Concepts, Machine Learning for Robot Control, Imitation Learning
\end{IEEEkeywords}

\IEEEpeerreviewmaketitle

\section{Introduction}
\IEEEPARstart{O}{ne} of the grand challenges in robotics is to create a generalist robot: a single agent capable of performing a wide variety of tasks in everyday settings based on arbitrary user commands. Doing so requires the robot to acquire a diverse repertoire of general-purpose skills and  non-expert users to be able to effectively specify tasks for the robot to solve.
This stands in contrast to most current end-to-end models,  which typically learn individual tasks one at a time from manually-specified rewards and assume tasks being specified via goal images~\cite{lynch2020learning} or one-hot skill selectors~\cite{ kalashnikov2021mt}, which are not practical for untrained users to instruct robots. 
Not only is this inefficient, but also limits the versatility and adaptivity of the systems that can  be  built.
How can we design learning systems that can efficiently acquire a diverse repertoire of useful skills that allows them to solve many different tasks based on arbitrary user commands?

To address this problem, we must resolve two questions. (1) How can untrained users direct the robot to perform specific tasks? Natural language presents a promising alternative form of specification, providing an intuitive and flexible way for humans to communicate tasks and refer to abstract concepts. However, learning to follow language instructions involves addressing a difficult symbol grounding problem~\cite{harnad1990symbol}, relating a language instruction to a robot's onboard perception and actions. (2) How can the robot efficiently learn general-purpose skills from offline data, without hand-specified rewards? 
A simple and versatile choice is to define skills as being continuous instead of discrete, endowing the agent of task-agnostic control: the ability to reach any reachable goal state from any current state~\cite{kaelbling1993learning}. These forms of task specification can in principle enable a robot to solve multi-stage tasks by following several language instructions in a row.
 \begin{figure}[t]
	\centering
	\includegraphics[width=1\columnwidth]{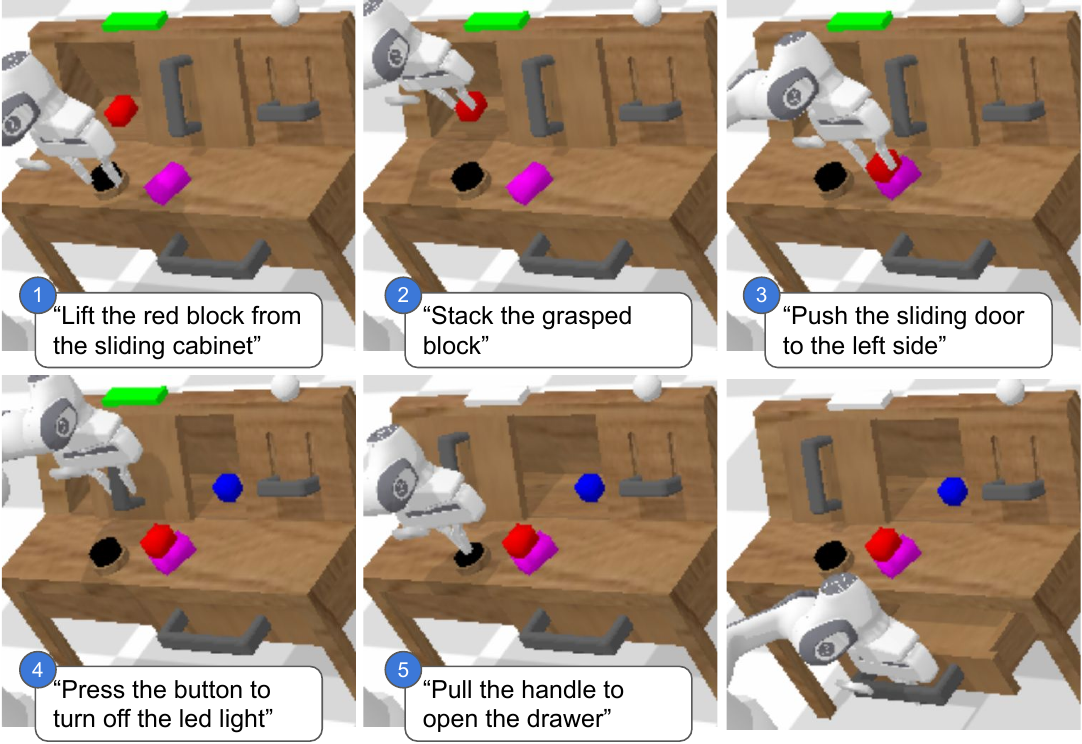}
	\caption{\gls{coolname} learns a single 7-DoF language conditioned visuomotor policy from offline, unstructured data that can solve multi-stage, long-horizon robot manipulation tasks. We divide instruction following into learning global plans representing high-level behavior and a local policy conditioned on the plan and the instruction.}
	\label{fig:cover_lady}
\end{figure}

Recent advances have been made at learning language conditioned  policies for continuous visuomotor-control in 3D environments via imitation learning~\cite{lynch2020language, stepputtis2020language, jang2021bc, team2021creating}  or reinforcement learning~\cite{nair2021learning, shaoconcept2robot}. These approaches typically require offline data sources of robotic interaction together with post-hoc crowd-sourced natural language labels. Although all methods share the basic idea of leveraging instructions that are grounded in the agent's high-dimensional observation space, their details vary greatly. Moreover, evaluating published methods and their components in language conditioned policy learning is difficult due to incomparable setups or subjective task definitions. 
In this work we systematically compare, improve, and integrate key components by leveraging the recently proposed CALVIN benchmark~\cite{calvin21} to further our understanding and provide a unified framework for long-horizon language conditioned policy learning. We build upon relabeled imitation learning~\cite{andrychowicz2017hindsight} to distill many reusable behaviors into a goal-directed policy, as seen in Fig.~\ref{fig:cover_lady}. Our approach consists of only standard supervised learning subroutines, and learns perceptual and linguistic understanding, together with task-agnostic control end-to-end as a single neural network.
Our contributions are:
\begin{itemize}
    \item We systematically compare key components of language conditioned imitation learning over unstructured data, such as observation and action spaces, losses for aligning visuo-lingual representations, language models and latent plan representations, and we analyze the effect of other choices, such as data augmentation and optimization.
    \item We propose four improvements to these key components: a multimodal transformer encoder to learn to recognize and organize behaviors during robotic interaction into a global categorical latent plan, a hierarchical division of the robot control learning that learns local policies in the gripper camera frame conditioned on the global plan, balancing terms within the KL loss and a self-supervised contrastive visual-language alignment loss.
    \item We integrate the best performing improved components in a unified framework, \acrfull{coolname}. Our model sets a new state of the art on the challenging CALVIN benchmark~\cite{calvin21}, on learning a single 7-DoF policy that can perform  long-horizon manipulation tasks in a 3D environment, directly from images, and only specified with natural language.
\end{itemize}

\section{Related Work}
Natural language processing has recently received much attention in the field of robotics~\cite{tellex2020robots}, following the advances made towards learning groundings between vision and language~\cite{lu2019vilbert, radford2021learning} and grounding behaviors in language~\cite{winograd1972understanding}. Early works have approached instruction following by designing interactive fetching systems to localize objects mentioned
in referring expressions~\cite{Shridhar-RSS-18, hatori2018interactively} or by grounding not only objects, but also spatial relations to follow language expressions characterizing pick-and-place commands~\cite{mees21iser, liu2021structformer, shridhar2021cliport}.
Unlike these approaches, we directly learn robotic control from images and natural language instructions, and do not assume any predefined motion primitives.

More recently, end-to-end deep learning has been used to condition agents on natural language instructions~\cite{lynch2020language, stepputtis2020language, jang2021bc, team2021creating, nair2021learning, shaoconcept2robot}, which are then trained under an imitation or reinforcement learning objective.
These works have pushed the state of the art and generated a range of ideas for language conditioned policy learning, such as losses for aligning visual observations and language instructions. However, each work evaluates a different combination of ideas and uses different setups or task definitions, making it unclear how individual ideas compare to each other and which ideas combine well together.
For example, the methods BC-Z and MIA~\cite{jang2021bc, team2021creating} use both behavior cloning, but different actions spaces and multi-modal alignment losses, such as regressing the language embedding from visual observations~\cite{jang2021bc} or cross-modality matching~\cite{team2021creating}. Moreover, BC-Z leverages expert trajectories and task labels, and MIA includes mobile navigation, making them difficult to implement directly in CALVIN, which contains unlabeled play data on different tabletop environments. Nair \emph{et. al.}~\cite{nair2021learning} learn a reward classifier which predicts if a change in state completes
a language instruction and leverage it for offline multi-task RL given four camera views. Similar to BC-Z they rely on discrete task labels and do not focus on solving long-horizon language-specified tasks. Most related to our approach is multi-context imitiation learning (MCIL)~\cite{lynch2020language}, which also uses relabeled  imitation  learning  to  distill reusable behaviors into a goal-reaching policy. 
Besides different action and observation spaces, these works leverage different language models to encode the raw text instructions into a semantic pre-trained vector space, making it difficult to analyze which language models are best suited for language conditioned policy learning.
The ablation studies presented in these papers show that each novel contribution of each work does indeed improve the performance of their model, but due to incomparable setups and evaluation protocols, it is difficult to asses what matters for language conditioned policy learning. 
Our work addresses this problem by systematically comparing and combining different observation and action spaces, auxiliary losses and latent representations and integrating the best performing components in a unified framework.

\section{Problem Formulation and Method Overview}
We consider the problem of learning a goal-conditioned policy  $ \pi_{\theta} \left(a_t \mid s_t, l \right)$ that outputs action $a_t \in \mathcal{A}$, conditioned on the current state $s_t \in \mathcal{S}$ and free-form language instruction $l \in \mathcal{L}$, under environment dynamics $ \mathcal{T}: \mathcal{S} \times \mathcal{A} \rightarrow \mathcal{S}$. We note that the agent does not have access to the true state of the environment, but to visual observations.
In CALVIN~\cite{calvin21} the action space $\mathcal{A}$ consists of the 7-DoF  control of a Franka Emika  Panda robot arm  with a parallel gripper.

We model the interactive agent with a general-purpose goal-reaching policy based on multi-context imitation learning (MCIL) from play data~\cite{lynch2020language}. To learn from unstructured ``play'' we assume access to an unsegmented teleoperated play dataset $\mathcal{D}$ of semantically meaningful behaviors provided by users, without a set of predefined tasks in mind. 
To learn control, this long temporal state-action stream $\mathcal{D} = \{ (s_t, a_t )\}_{t=0}^{\infty}$ is relabeled~\cite{andrychowicz2017hindsight}, treating each visited state in the dataset as a ``reached goal state'', with the preceding states and actions treated as optimal behavior for reaching that goal. Relabeling yields a dataset of $D_{\text{play}} = \{ (\tau, s_g )_i\}_{i=0}^{D_{\text{play}}}$ where each goal state $s_g$ has a trajectory demonstration $\tau=\{ (s_0, a_0),\ldots \}$ solving for the goal. These short horizon goal image conditioned demonstrations can be fed to a simple maximum likelihood goal conditioned imitation objective:
\begin{equation}
    \mathcal{L}_{LfP} = \mathbb{E}_{(\tau, s_g) \sim D_{\text{play}}}  \left [ \sum_{t=0}^{\mid \tau \mid} \log \pi_{\theta} (a_t \mid s_t, s_g)\right ] 
\end{equation}
to learn a  goal-reaching policy $\pi_{\theta} \left(a_t \mid s_t, s_g \right)$. We address the inherent multi-modality in free-form imitation datasets by auto-encoding contextual demonstrations through a latent ``plan'' space with an sequence-to-sequence conditional variational auto-encoder (seq2seq CVAE)~\cite{lynch2020learning}. Conditioning the policy on the latent plan frees up the policy to use the entirety of its capacity for learning uni-modal behavior.  To generate latent plans $z$ we make use of the variational inference framework~\cite{kingma2013auto}. The objective of the latent plan sampler is to model the full distribution over all high-level behaviors that might connect the current and goal state, to provide multi-modal plans at inference time.
This distribution is learned with a CVAE by maximizing the marginal log likelihood of the observed behaviors in the dataset $\log p(x \mid s)$, where $x$ are sampled state-action trajectories from $\tau$. The Evidence Lower Bound (ELBO)~\cite{kingma2013auto} for the CVAE can be written as:
\begin{equation}
    \log p(x|s)\geq -\textrm{KL}\big(q (z|x,s) ~||~ p(z|s)\big) + \mathbb{E}_{q(z|x,s)}\left[ \log p(x|z,s) \right]
\end{equation}
The decoder is a policy trained to reconstruct input actions, conditioned on state $s_t$, goal $s_g$, and an inferred plan $z$ for how to get from $s_t$ to $s_g$. At test time, it takes a goal as input, and infers and follows plan $z$  in closed-loop.

However, when learning language conditioned policies $ \pi_{\theta} \left(a_t \mid s_t, l \right)$ it is not possible to  relabel any visited state $s$ to a natural language goal as the goal space  is no longer equivalent to the observation space. Lynch \emph{et al.}~\cite{lynch2020language} showed that pairing a small number of random windows with language after-the-fact instructions enables learning a single
language conditioned visuomotor policy that can perform a wide variety of robotic manipulation tasks. The key insight here is that solving a single imitation learning policy for either goal image or language goals, allows for learning control mostly from unlabeled play data and reduces the burden of language annotation to less than 1\% of the total data. Concretely, given multiple contextual imitation datasets $\mathcal{D} = \{D^0, D^1,\ldots ,D^K\}$, with a different way of describing
tasks,  MCIL trains a single latent goal conditioned policy $\pi_{\theta} \left(a_t \mid s_t, z \right)$ over all datasets simultaneously, as well as one parameterized encoder per dataset.

 \begin{figure*}[t]
	\centering
	\includegraphics[scale=0.8]{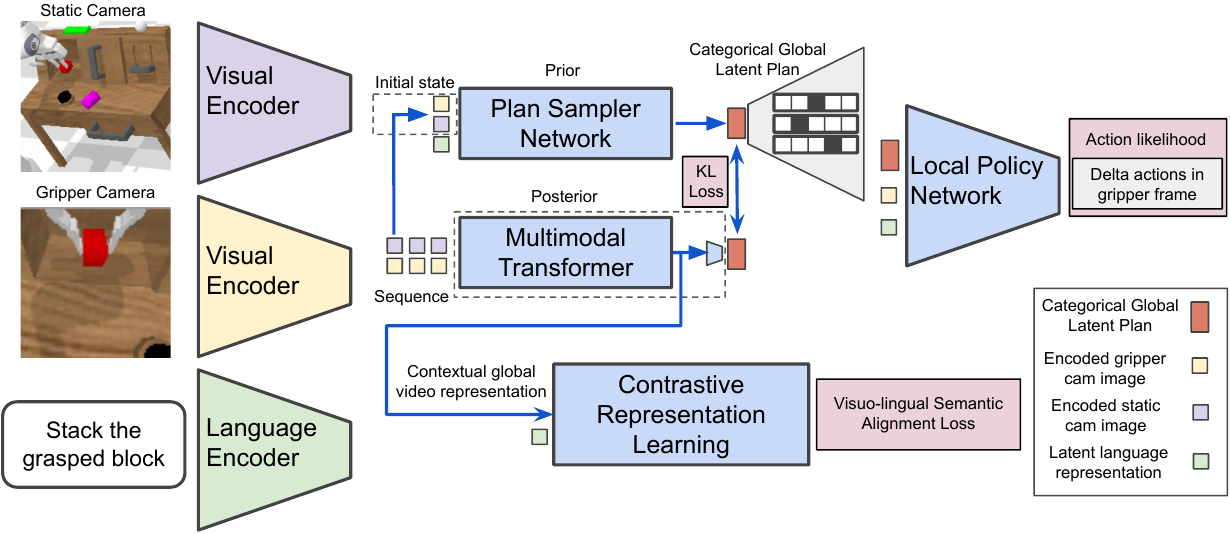}
	\caption{ Overview of our architecture to learn language conditioned policies from unstructured data. First the language instructions and the visual observations are encoded. During training a multimodal transformer encodes sequences of observations to learn to recognize and organize high-level behaviors through a posterior. Its temporally contextualized features are provided as input to a contrastive visuo-lingual alignment loss. The plan sampler network receives the initial state and the latent language goal and predicts the distribution over plans for achieving the goal. Both prior and posterior distributions are predicted as a vector of multiple categorical variables and are trained by minimizing their KL divergence. The local policy network receives the latent language instruction, the gripper camera observation and the global latent plan to generate a sequence of relative actions in the gripper camera frame to achieve the goal.}
	\label{fig:architecture}
\end{figure*}

\section{Key Components of Language Conditioned Imitation Learning over Unstructured Data}
This section compares and improves key components of language conditioned imitation learning over unstructured data.
We base our model on MCIL~\cite{lynch2020language} and improve it by decomposing control into a hierarchical approach of generating global plans with a static camera and learning local policies with a gripper camera conditioned on the plan. Then we go through different components that have a large impact on performance: architectures to encode sequences in relabeled imitation learning,
the representation of the latent distributions, how to best align language and visual representations, data augmentation and optimization.
We visualize the full architecture in Fig.~\ref{fig:architecture}.

\subsection{Observation and Actions Spaces}
How to best represent motion skills is an age-old question in robotics. From a learning perspective, generating the action sequences to solve diverse manipulation tasks with a single network from high-dimensional observations is challenging, because the distribution is multi-modal, discontinuous and imbalanced.
For these reasons, finding an efficient representation is crucial to perform this non-trivial reasoning using learning-based methods. MCIL~\cite{lynch2020language} uses global actions learned from a single static RGB camera. We observe that predicting 7-DoF global actions leads to the network primarily solving static element tasks, such as pushing a button, but failing to generalize to dynamic tasks, such as manipulating colored blocks. To alleviate this problem, we propose generating global plans  that correspond to reusable common behavior $b$ seen in the play data, but learning local policies conditioned on the plan. This results in a hierarchical approach that frees up the network from having to memorize all locations in the scene were the behaviors were performed. Concretely, we encode RGB images from both the static and a gripper camera to learn a compact representation of all the different high-level plans that take an agent from a current state to a goal state, learning $ p \left(b \mid s_t, s_g \right )$. Inspired by a recent line of work that aims to learn hierarchies of controllers based on static and gripper cameras~\cite{borja22icra}, we use the encoded gripper camera representations in the policy network, the global contextualized latent plan, and perform control in the gripper frame with relative actions for an efficient robot control learning. The action space consists of delta XYZ position, delta euler angles and the gripper action. Our proposed formulation has several advantages: a) local policies based on the gripper camera generalize better to different locations of the objects to be manipulated b) the policy has a prior in the form of a global contextualized latent plan, but is free to discover the exact strategy on how to interact with the objects.

\subsection{Latent Plan Encoding}
 A challenge in self-supervising control on top of free-form imitation data is that in general, there are many valid high-level behaviors that might connect the same $(s_t, s_g)$ pairs. By auto-encoding contextual demonstrations through a latent ``plan'' space with an sequence-to-sequence conditional variational auto-encoder (seq2seq CVAE)~\cite{lynch2020learning}, 
we can learn to recognize which region of the latent plan space an observation-action sequence belongs to. Critically, conditioning the policy on the latent plan frees up the policy to use the entirety of its capacity for learning uni-modal behavior. Thus, learning to generate and represent high-quality latent plans is a key component in the seq2seq CVAE framework. MCIL~\cite{lynch2020language} uses bidirectional recurrent neural networks (RNN) to encode a randomly sampled play sequence and map it into a latent Gaussian distribution. In contrast, we leverage a multimodal transformer encoder~\cite{vaswani2017attention} to build a contextualized representation of abstract behavior 
 expressed in language instructions and map into a vector of several latent categorical variables~\cite{hafner2020mastering}.
 The foundation of the  Transformer architecture is the scaled dot-product attention function, which enables elements in a sequence to attend to other elements. The attention function receives as input a sequence $\{x_1,...,x_n\}$ and outputs a sequence $\{y_1,...,y_n\}$. Each input $x_i$ is projected linearly to a query $q_i$, key $k_i$, and value $v_i$. To compute the output $y_i$ the values are summed with weights that take into account the similarity of the query with its corresponding key.
 The attention function is defined as 
$\text{Attention}(Q,K,V) = \text{softmax}(\frac{QK^T}{\sqrt{d_k}})V$, where $d_k$ is the dimension of the keys and queries. The queries, keys, and values are stacked together into matrix $Q \in \mathbb{R}^{n \times d_{\text{model}}}$, $K \in \mathbb{R}^{n \times d_{\text{model}}}$, and $V \in \mathbb{R}^{n \times d_{\text{model}}}$.
 We encode the sequence of visual observations of both modalities  $X_{\{static, gripper\}} \in \mathbb{R}^{T \times H \times W \times 3}$ with separate perceptual encoders, and concatenate them to form the fused perceptual representation $V \in \mathbb{R}^{T \times d}$ of the sampled demonstration, where $T$ represents the sequence length and $d$ the feature dimension. To enable the sequences to carry temporal information, we add positional embeddings~\cite{vaswani2017attention} and feed the result into the Multimodal Transformer 
 to learn temporally contextualized global video representations. Finally, inspired by the recent line of work that looks into learning discrete instead of continuous latent codes~\cite{van2017neural, hafner2020mastering}, we represent the latent plans as a vector of multiple categorical latent variables and  and optimize them using straight-through gradients~\cite{bengio2013estimating}. 
 Learning discrete representations in the context of language conditioned policies is a natural fit, as language is inherently discrete and images can often be described concisely by language~\cite{mees21iser}. Furthermore, discrete representations are a natural fit for complex reasoning, planning and predictive learning (e.g., if it is sunny, I will go to the beach).

\subsection{Semantic Alignment of Video and Language}
Learning to follow language instructions involves addressing a difficult symbol grounding problem~\cite{harnad1990symbol}, relating a language instruction to a robots onboard perception and actions. Although instructions and visual observations are aligned in CALVIN, learning to manipulate the colored blocks is a challenging problem. This is due to the fact that the robot needs to learn a wide variety of diverse behaviors to manipulate the blocks, but also needs to understand which colored block the user is referring to. Thus, the block related instructions are very similar, for the exception of a word that might disambiguate the instruction by indicating a color. Therefore, most pre-trained language models struggle to learn such semantics from text only and the policy needs to learn referring expression comprehension via the imitation loss.
There have been a number of multi-modal alignment losses proposed, such as regressing the language embedding from the visual observation~\cite{jang2021bc} or cross-modality matching~\cite{team2021creating}.
 We maximize the cosine similarity between the visual features of the sequence $i$ and the corresponding language features while, at the same time, minimizing the cosine similarity between the current visual features and other language instructions in the same batch.
 We define our $ \mathcal{L}_{contrast}$ loss the same way as the contrastive loss for pairing images and captions in CLIP~\cite{radford2021learning}. However, ideally our model should use the time-dependent representation of the sequence visual observations in order to capture the meaning of a language instruction. This can be appreciated only after the sequence of actions have been executed for several timesteps. The usage of in-batch negatives enables re-use of computation both in the forward and the backward pass making training highly efficient. The logits for one batch is a $M\times M$ matrix, where each entry is given by $\text{logit}(x_i, y_j)  = \text{cos\_sim}(x_i, y_j) \cdot \exp(\tau),\forall (i,j), i,j \in \{1,2,\ldots,M\}$
where $\tau$ is a trainable temperature parameter. Only entries on the diagonal of the matrix are considered positive examples. The final loss is the sum of the cross entropy losses on the row and the column direction.

\subsection{Action Decoder}
A challenge in learning control from free-form imitation data, in which different ways of executing the same skill are shown, is that a standard unimodal predictor, such as a Gaussian distribution, will average out dissimilar motions. To address this multimodality, we follow the solution proposed by Lynch \emph{et. al.}~\cite{lynch2020learning} of discretizing the action space and then parameterizing the policy as a discretized logistic mixture distribution~\cite{salimans2017pixelcnn++, dasari2021transformers}. 
Each of the predicted $k$ logistic distributions have a separate mean and scale, and are weighed with $\alpha$ to form the mixture distribution. 
The imitation loss is the negative log-likelihood for this distribution:
$$ \mathcal{L}_{act}(\mathcal{D}_{play}, V) = - \ln(\Sigma_{i=0}^k \hspace{1mm} \alpha_k(V_t) \hspace{1mm} P(a_t, \mu_i(V_t), \sigma_i(V_t))$$
Where, $P(a_t, \mu_i(V_t), \sigma_i(V_t)) = F(\frac{a_t + 0.5 - \mu_i(V_t)}{\sigma_i(V_t)}) - F(\frac{a_t - 0.5 - \mu_i(V_t)}{\sigma_i(V_t)})$ and $F(\cdot)$ is the logistic CDF. Additionally, we use a cross-entropy loss to model the binary gripper open/close action.

\subsection{Optimization and Implementation Details}
Our full training objective for the  1\% of the total data that is annotated with after-the-fact language instructions is given by $ \mathcal{L} =  \mathcal{L}_{act} + \beta  \mathcal{L}_{KL} +  \lambda \mathcal{L}_{contrast}$. The windows without annotations are trained with the same imitation learning objective, but the language goals are replaced by the last visual frame of the sampled window to learn control in a fully self-supervised manner. A common problem in training VAEs is finding the right balance in the weight of the KL loss. A high $\beta$ value can result in an over-regularized model in which the decoder ignores the latent plans from the prior, also known as a ``posterior collapse''~\cite{bowman2015generating}. On the other hand, setting $\beta$ too low results in the plan sampler network being unable to catch up to plan over the latent space created by the posterior, and as a result at test time the plans generated by the plan sampler network will be unfamiliar inputs for the decoder. Orthogonal to this, as the KL loss is bidirectional, we want to avoid regularizing the plans generated by the posterior toward a poorly trained prior. To solve this problem, we minimize the KL loss
faster with respect to the prior than the posterior by using different learning rates, $\alpha = 0.8$ for the prior and $1 - \alpha$ for the  posterior, similar to Hafner \emph{et. al.}~\cite{hafner2020mastering}. We set $\beta = 0.01$ and $\lambda = 3$ for all experiments and train with the Adam optimizer with a learning rate of $2^{-4}$. During training, we randomly sample windows between length 20 and 32 and pad them until the max length of 32. For the latent plan representation we use 32 categoricals with 32 classes each. To better compare the differences  between approaches, we use the same convolutional encoders as the MCIL baseline available in CALVIN for processing the images of the static and gripper camera.
Our multimodal transformer encoder has 2 blocks, 8 self-attention heads, and a hidden size of 2048. In order to encode raw text into a semantic pre-trained vector space, we leverage the paraphrase-MiniLM-L3-v2 model~\cite{reimers-2019-sentence-bert}, which distills a large Transformer based language model and is trained on paraphrase language corpora that is mainly derived from Wikipedia. It has a vocabulary size of 30,522 words and maps a sentence of any length into a vector of size 384.

\subsection{Data Augmentation}
To aid learning we apply data augmentation to image observations, both in our method and across all baselines.  During training, we apply stochastic image shifts of 0-4 pixels to the gripper camera images and of 0-10 pixels to the static camera images as in Yarats \emph{et. al.}~\cite{yarats2021drqv2}. Additionally, a bilinear interpolation is applied
 on top of the shifted image by replacing each pixel with the average of the nearest pixels.

\section{Experiments}
We evaluate our model in an extensive comparison and ablation study, to determine which components matter for language conditioned imitation learning over unstructured data. 
We ablate single components of our full approach to study the influence of each component. 
We then compare our resulting model to the best published methods on the CALVIN benchmark, and show that it outperforms all previous methods.

 \begin{figure*}[h]
  \centering
  \begin{tabular}{|c |c| c| c| c| c|  c| }
  \hline
  Method  & \multicolumn{6}{|c|}{LH-MTLC}\\
  \hline
    &     \multicolumn{6}{|c|}{No. Instructions in a Row (1000  chains)}\\
        \cline{2-7}
&          1 & 2 & 3 & 4 & 5 & Avg. Len.\\
  \hline
 MCIL~\cite{lynch2020language}     & 34.4\% & 5.8\% & 1.1\% & 0.2\% & 0.08\% & 0.41\\
 GCBC~\cite{lynch2020learning} + delta actions      & 64.7\% (4.0) & 28.4\% (6.2) & 12.2\% (4.1) & 4.9\% (2.0) & 1.3\% (0.9) & 1.11 (0.3)\\
 MCIL~\cite{lynch2020language} + delta actions      & 76.4\% (1.5) & 48.8\% (4.1) & 30.1\% (4.5) & 18.1\% (3.0) & 9.3\% (3.5) & 1.82 (0.2)\\
  Ours        & \textbf{82.7}\% (0.3)  & \textbf{64.9}\% (1.7) & \textbf{50.4}\% (1.5) & \textbf{38.5}\% (1.9)& \textbf{28.3}\% (1.8) & \textbf{2.64} (0.05)\\
  \hline
  No Transformer Encoder         & 79.5\% (0.6) & 61.5\% (2.0) & 46.7\% (1.7) & 32.6\% (1.2)& 24.7\% (1.7) & 2.45 (0.06)\\
  No Discrete Latents         & 79.8\% (1.6)  & 60.6\% (0.4) & 43.3\% (0.9) & 32.6\% (1.1) & 23.6\% (1.6) & 2.39 (0.03)\\
  No Local Policy       & 78.4\% (1.4) & 56.2\% (2.0) & 40.4\% (2.4) & 29.5\% (1.7)  & 20.1\% (1.3) & 2.24 (0.06)\\
  No KL Balancing         & 79.6\% (2.3)  & 59.3\% (0.2) & 42.3\% (0.7) & 30.7\% (0.8) & 21.9\% (1.1) & 2.33 (0)\\
  No Gripper Log Loss         & 79.5\% (1.3)  & 61.3\% (1.7) & 46.5\% (2.1) & 33.9\% (1.9) & 24.0\% (1.8) & 2.45 (0.07)\\
  KL $\beta=0.1$         & 78.4\% (1.5) & 55.8\% (1.4) & 36.3\% (1.7) & 23.9\% (1.1) & 16.2\% (1.2) & 2.10 (0.09)\\

  \rowcolor{Gray}
  No Lang Align. Loss         & 80.1\% (0.9) & 55.4\% (2.0) & 42.2\% (1.3) & 30.2\% (1.3) & 21.8\% (0.9) & 2.29 (0.07)\\
  \rowcolor{Gray}
  MIA Lang Align. Loss~\cite{team2021creating}         & 79.5\% (1.9)  & 57.8\% (0.8) & 41.7\% (1.4) & 29.7\% (2.1) & 20.9\% (1.6) & 2.29 (0.04)\\
  \rowcolor{Gray}
  Lang Align. Regression~\cite{jang2021bc}         & 82.5\% (0.8)  & 61.0\% (1.0) & 45.5\% (0.4) & 32.9\% (1.1) & 23.5\% (0.7) & 2.45 (0.03)\\
  No image augmentation        & 75.6\% (1.4) & 51.2\% (1.7) & 34.0\% (2.6) & 23.6\% (1.9) & 15.4\% (0.6)& 1.99 (0.07)\\
  Proprioceptive input         & 81.2\% (0.3) & 56.1\% (1.4) & 39.2\% (1.6) & 26.2\% (1.3) & 18.1\% (0.8) & 2.20 (0.05)\\
  
  \rowcolor{Gray}
  MPNet SBERT~\cite{reimers-2019-sentence-bert}        & 77.4\% (2.0)  & 57.2\% (1.9) & 40.7\% (1.4) & 28.5\% (1.7) & 20.2\% (1.0) & 2.24 (0.06)\\
  \rowcolor{Gray}
  MPNet~\cite{song2020mpnet}          & 71.5\% (2.6) & 48.5\% (1.4) & 33.6\% (1.6)& 23.1\% (1.7) & 15.5\% (0.7) & 1.92 (0.08)\\
  \rowcolor{Gray}
  Distilroberta SBERT~\cite{reimers-2019-sentence-bert}       & 78.1\%  (1.2) & 60.9\% (1.5) & 47.7\% (1.9) & 36.4\% (1.9) & 27.5\% (0.7) & 2.50 (0.07)\\
  \rowcolor{Gray}
  Distilroberta~\cite{liu2019roberta}        & 77.8\% (1.4) & 56.3\% (1.7)& 40.6\% (2.4) & 28.2\% (1.7) & 17.8\% (1.9) & 2.21 (0.07)\\
  \rowcolor{Gray}
  BERT~\cite{devlin2018bert}        & 77.5\% (1.8)  & 52.6\% (3.2) & 34.7\% (2.4)& 24.1\% (1.5) & 14.9\% (2.4) & 2.03 (0.1)\\
  CLIP Encoders~\cite{radford2021learning}       & 81.4\% (0.5)  & 60.4\% (1.1) & 44.7\% (2.3) & 32.3\% (1.2) & 23.2\% (1.6)& 2.42 (0.06)\\

  \hline
  \end{tabular}
  \caption{Performance of our model on the D environment of the CALVIN Challenge and ablation of the key components, across 3 seeded runs. All models receive RGB images from both a static and a gripper camera as a input.}
  \label{tab:roboexp}
\end{figure*}

\subsection{Evaluation Protocol}
The goal of the agent in CALVIN is to solve sequences of up to 5 language instructions in a row using only onboard sensors. This setting is very challenging as it requires agents to be able to transition between different subgoals. CALVIN has a total of 34 different subtasks and evaluates 1000 unique sequence instruction chains.
The robot is set to a neutral position after every sequence to avoid biasing the policies through the robot's initial pose. This neutral initialization breaks correlation between initial state and task, forcing the agent to rely entirely on language to infer and solve the task. For each subtask in a row the policy is conditioned on the current subgoal instruction and transitions to the next subgoal only if the agent successfully completes the current task. We perform the ablation studies on the environment D of CALVIN and additionally report numbers of our approach for the other two CALVIN splits, the multi environment and zero-shot multi environment splits. We emphasize that the CALVIN dataset for each of the four environment consists of 6 hours of teleoperated undirected play data that might contain suboptimal behavior. To simulate a real-world scenario, only 1\% of that data contains crow-sourced language annotations.
\subsection{Results and Ablations of Key Components}
\textit{Observation and Actions Spaces}: We compare our approach of dividing the robot control learning into generating global contextualized plans and conditioning a local policy that receives only the observations of the the gripper camera on the global plan against a ``No Local Policy'' baseline. Unlike our approach, which performs control in the gripper camera frame, the baseline's policy receives both cameras images and performs control in the robot's base frame, as is usual in most published approaches. We observe in Fig.~\ref{tab:roboexp}, that despite the baseline's decoder having more perceptual information, the performance for completing 5 chains of language instructions sequentially drops from 28.3\% to 20.1\%. 
In order to analyze the big performance difference with respect to the original MCIL baseline, we train a MCIL baseline with relative actions and observe that its performance improves significantly from the original MCIL baseline with absolute actions, but performs worse than our models. We speculate that using relative actions with a local policy is easier for the agent to learn instead of memorizing all the locations where interactions have been performed with global actions and a global observation space. By decoupling the control into a hierarchical structure, we show that performance increases significantly. Additionally, we analyze the influence of using the 7-DoF proprioceptive information as input for both the plan encodings and conditioning the policy, as  many works report improved performance from it~\cite{lynch2020learning, lynch2020language, kalashnikov2021mt}. We observe that the performance drops significantly and the agent relies too much on the robot's initial position, rather than learning to disentangle initial states and tasks. We hypothesize this might be due to a causal confusion between the proprioceptive information and
the target actions~\cite{de2019causal}. We also analyze the effect of modeling the full action space, including the binary gripper action dimension, with the mixture of logistics distribution instead of using the log loss for the open/close gripper action and observe that the average sequence length drops from 2.64 to 2.45. Finally, we note that applying stochastic image shifts to the input images increases the performance significantly.
\begin{figure*}[ht]
  \centering
  \begin{tabular}{|c | c| c| c|  c| c| c |c |}
  \hline
  Method  & Train$\,\to\,$Test & \multicolumn{6}{|c|}{LH-MTLC}\\
   \hline
   &     & \multicolumn{6}{|c|}{No. Instructions in a Row (1000  chains)}\\
\cline{3-8}
 &     &     1 & 2 & 3 & 4 & 5 & Avg. Len.\\
   \hline
  MCIL~\cite{lynch2020language} & A,B,C,D$\,\to\,$D  & 37.3\% & 2.7\% & 0.17\% & 0\% & 0\% & 0.40\\
  Ours & A,B,C,D$\,\to\,$D  & \textbf{88.9}\% (0.6) & \textbf{73.3}\% (1.7) & \textbf{58.7}\% (1.8) & \textbf{47.5}\% (1.6) & \textbf{38.3}\% (1.9) & \textbf{3.06} (0.07)\\
  \hline
  MCIL~\cite{lynch2020language} & A,B,C$\,\to\,$D    & 30.4\% & 1.3\% & 0.17\% & 0\% & 0\% & 0.31\\
  Ours & A,B,C$\,\to\,$D  & \textbf{41.8}\% (2.3) & \textbf{16.5}\% (2.5) & \textbf{5.7}\% (1.3) & \textbf{1.9}\% (0.9) & \textbf{1.1}\% (0.5) & \textbf{0.67} (0.1)\\
\hline
  \end{tabular}
  \caption{Performance of our model on the multi environment splits of the CALVIN Challenge across 3 seeded runs. }\label{tab:multienv}
\end{figure*}

\begin{figure}[h]
	\centering
	\includegraphics[width=1\columnwidth]{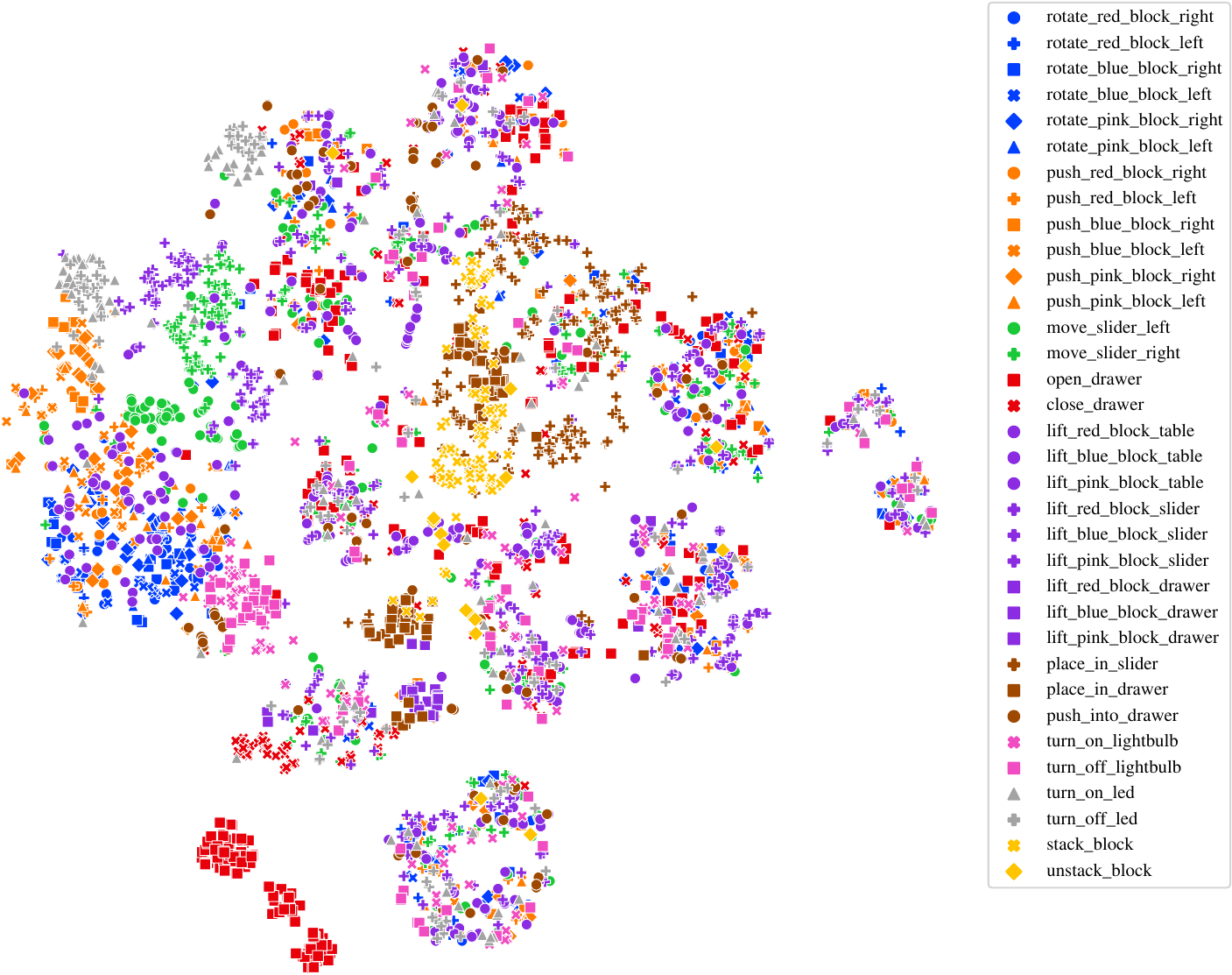}
	\caption{t-SNE visualization of the discrete latent plans generated by embedding randomly selected unseen language annotations.
	Surprisingly, we find that despite not being trained explicitly with task labels, \gls{coolname} appears to organize its latent plan space functionally. We visualize with the same color functionally similar skills, but use different shapes to distinguish sub-skills.}
	\label{fig:tsne}
\end{figure}

\textit{Latent Plan Encoding}: In our CVAE framework the latent plan represents valid ways of connecting the actual state and the goal state and thus, frees up the policy to use the entirety of its capacity for learning uni-modal behavior. As language is inherently discrete and discrete representations are a natural fit for complex reasoning and planning, we represent 
latent plans as a vector of multiple categorical latent variables and  and optimize them using straight-through gradients~\cite{bengio2013estimating}. We observe that the performance for 5-chain evaluation drops from 28.3\% to 23.6\% when we train our model with a diagonal Gaussian distribution as in MCIL.  While it is difficult to judge why categorical latents work better than 
 continuous latent variables, we hypothesize that categorical latents could  be a better inductive bias for non-smooth aspects of the CALVIN benchmark, such as when a block is hidden behind the sliding door. Besides, the sparsity level enforced by a categorical distribution could be beneficial for generalization. 
 Additionally, we compare against a goal-conditioned Behavior Cloning (GCBC) baseline~\cite{lynch2020learning} which does not condition the policy on a latent plan, and observe that it performs worse than MCIL with relative actions, highlighting the importance of modeling latent behaviors in free-form imitation datasets.
We also observe that balancing the KL loss is beneficial in the CVAE training.  By scaling up the prior cross entropy relative to the posterior entropy, the agent is encouraged to minimize the KL loss by improving its prior toward the more informed posterior, as opposed to reducing the KL by increasing the posterior entropy.
We visualize a t-SNE plot of our learned discrete latent space in Figure~\ref{fig:tsne} and that see that even for unseen language instructions it appears to organize the latent space functionally. Additionally, we report degraded performance for an over-regularized model which learns to ignore the latent plans, in which we weight the KL divergence with $\beta=0.1$. Finally, we evaluate replacing the transformer encoder in the posterior with a GRU bidirectional recurrent network of the same hidden dimension of 2048, similar to MCIL. The results suggest 
that besides an improved performance, the multimodal transformer encoder is significantly more efficient both memory and model size wise (5.9 M vs 106 M parameters for the posterior network) and overall training wall clock time. For comparison, with the transformer encoder, our full approach contains 47.1 M trainable parameters.

\textit{Semantic Alignment of Video and Language}:
One of the main challenges for language conditioned continuous visuomotor-control is solving a difficult symbol grounding problem~\cite{harnad1990symbol}, relating a language instruction to a robots onboard perception and actions.
An agent in CALVIN needs to learn a wide variety of diverse behaviors to manipulate blocks with different shapes, but also needs to understand which colored block the user is instructing it to manipulate.
We compare commonly used auxiliary losses for aligning visual and language representations. Concretely, we compare our contrastive loss against predicting the language embedding from the sequence's visual observations with a cosine loss~\cite{jang2021bc}, cross-modality matching~\cite{team2021creating} and not having an auxiliary visuo-lingual alignment loss.
We observe that using an auxiliary loss to semantically align the sampled video sequences and the language instructions helps, but both baselines perform similarly. We hypothesize that our contrastive loss works best because it leverages a larger number of in-batch negatives than the cross-modality matching loss. 
Concretely, we maximize the cosine similarity for $N$ real pairs in the batch while minimizing the cosine similarity of the multimodal embeddings of the $N^2 - N$ incorrect pairings. The cross-modality matching loss implements a discriminator that produces a binary predictor of whether the
embeddings match or not. The batch is shuffled only once to produce the negative samples, contrasting only $N$ negative samples.

\textit{Language Models}: Despite steady progress in language conditioned policy learning, a fundamental, but less considered aspect is the choice of the pre-trained language model to encode raw text into a semantic pre-trained vector space. We compare the lightweight paraphrase-MiniLM-L3-v2 language embeddings from our full model against several popular alternatives, such as the larger BERT~\cite{devlin2018bert}, Distilroberta~\cite{liu2019roberta} and MPNet~\cite{song2020mpnet}, which double the embedding size from 384 to 768. Besides the architecture of the language model, we analyze the impact of the loss functions the language models are trained on, by comparing the original embeddings of MPNet and Distilroberta against versions that have been finetuned with contrastive losses at the sentence level to map semantically similar sentences into the same latent space~\cite{reimers-2019-sentence-bert}. We observe that the SBERT models that have been finetuned on sentence semantic similarity achieve significantly better results than the original language models trained on masked language modeling. Concretely, the original Distilroberta model achieves an average sequence length of 2.21, while the SBERT Distilroberta model achieves an average sequence length of 2.50. Finally, we also compare against a model conditioned on visual (ResNet-50) and language-goal features from a pre-trained CLIP model~\cite{radford2021learning}, which has been trained to align visual and language features from millions of image-caption pairs from the internet. Surprisingly, we find that performance is slightly worse than our best performing model. We hypothesize that this might be due to a domain gap between the natural images that CLIP has been trained on and the simulated images from CALVIN.
The results suggest that for complex semantics, the choice of the pre-trained language model has a large impact and models finetuned on sentence level semantic similarity should be preferred.
While in this paper, we do not finetune the language models with the action loss, we anticipate this might lead to better performance, specially in order to ground instructions referring to the colored blocks.

\textit{Multi Environment and Zero-Shot Generalization}: Finally, we investigate the performance of our approach on the larger multi environment splits of CALVIN on Fig.~\ref{tab:multienv}. On the zero-shot split, which consists on training on three environments and testing on an unseen environment with unseen instructions, we observe that despite modest improvements over the MCIL baseline, the policy achieves just an average sequence length of 0.67. We hypothesize that in order to achieve better zero-shot performance,  additional techniques from the domain adaptation literature, such as adversarial skill-transfer losses might be helpful\cite{mees20icra_asn}. On the split that trains on all four environments and evaluates on one of them, we observe that \gls{coolname} benefits from the larger dataset size and sets a new state of the art with an average sequence length of 3.06, which is higher than our best performing model trained and tested on environment D (2.64). The results suggest that increasing the number of collected language pairs aids addressing the complicated perceptual grounding problem.

\section{Conclusion}
\label{sec:conclusion}
We have presented a study into what matters in language conditioned robotic imitation learning over unstructured data that systematically analyzes, compares, and improves a set of key components.
This study results in a range of novel observations about these components and their interactions, from which we integrate the best components and improvements into a state-of-the-art approach. 
Our resulting hierarchical \gls{coolname} model learns a single policy from unstructured imitation data that substantially surpasses the state of the art on the challenging language conditioned long-horizon robot manipulation CALVIN benchmark.
We hope it will be useful as a starting point for further research and will bring us closer towards general-purpose robots that can relate human language to their perception and actions.

\section*{Acknowledgments}
This work has  been supported partly by the German Federal Ministry of Education and Research under contract 01IS18040B-OML.

\bibliographystyle{IEEEtran}
\bibliography{references}

\begin{thebibliography}{10}
\providecommand{\url}[1]{#1}
\csname url@rmstyle\endcsname
\providecommand{\newblock}{\relax}
\providecommand{\bibinfo}[2]{#2}
\providecommand\BIBentrySTDinterwordspacing{\spaceskip=0pt\relax}
\providecommand\BIBentryALTinterwordstretchfactor{4}
\providecommand\BIBentryALTinterwordspacing{\spaceskip=\fontdimen2\font plus
\BIBentryALTinterwordstretchfactor\fontdimen3\font minus
  \fontdimen4\font\relax}
\providecommand\BIBforeignlanguage[2]{{%
\expandafter\ifx\csname l@#1\endcsname\relax
\typeout{** WARNING: IEEEtran.bst: No hyphenation pattern has been}%
\typeout{** loaded for the language `#1'. Using the pattern for}%
\typeout{** the default language instead.}%
\else
\language=\csname l@#1\endcsname
\fi
#2}}

\bibitem{lynch2020learning}
C.~Lynch, M.~Khansari, T.~Xiao, V.~Kumar, J.~Tompson, S.~Levine, and
  P.~Sermanet, ``\href{https://arxiv.org/pdf/1903.01973.pdf}{Learning latent
  plans from play},'' in \emph{CoRL}, 2019.

\bibitem{kalashnikov2021mt}
D.~Kalashnikov, J.~Varley, Y.~Chebotar, B.~Swanson, R.~Jonschkowski, C.~Finn,
  S.~Levine, and K.~Hausman,
  ``\href{https://arxiv.org/pdf/2104.08212.pdf}{MT-Opt: Continuous Multi-Task
  Robotic Reinforcement Learning at Scale},'' \emph{arXiv preprint
  arXiv:2104.08212}, 2021.

\bibitem{harnad1990symbol}
S.~Harnad, ``The symbol grounding problem,'' \emph{Physica D: Nonlinear
  Phenomena}, vol.~42, no. 1-3, pp. 335--346, 1990.

\bibitem{kaelbling1993learning}
L.~P. Kaelbling, ``Learning to achieve goals,'' in \emph{IJCAI}, 1993, pp.
  1094--1099.

\bibitem{lynch2020language}
C.~Lynch and P.~Sermanet,
  ``\href{https://arxiv.org/pdf/2005.07648.pdf}{Language Conditioned Imitation
  Learning Over Unstructured Data},'' in \emph{RSS}, 2021.

\bibitem{stepputtis2020language}
S.~Stepputtis, J.~Campbell, M.~Phielipp, S.~Lee, C.~Baral, and H.~B. Amor,
  ``\href{https://proceedings.neurips.cc/paper/2020/file/9909794d52985cbc5d95c26e31125d1a-Paper.pdf}{Language-Conditioned
  Imitation Learning for Robot Manipulation Tasks},'' in \emph{NeurIPS}, 2020.

\bibitem{jang2021bc}
E.~Jang, A.~Irpan, M.~Khansari, D.~Kappler, F.~Ebert, C.~Lynch, S.~Levine, and
  C.~Finn,
  ``\href{https://proceedings.mlr.press/v164/jang22a/jang22a.pdf}{BC-0:
  Zero-Shot Task Generalization with Robotic Imitation Learning},'' in
  \emph{CoRL}, 2021.

\bibitem{team2021creating}
D.~I.~A. Team, J.~Abramson, A.~Ahuja, A.~Brussee, F.~Carnevale, M.~Cassin,
  F.~Fischer, P.~Georgiev, A.~Goldin, T.~Harley, \emph{et~al.},
  ``\href{https://arxiv.org/pdf/2112.03763.pdf}{Creating Multimodal Interactive
  Agents with Imitation and Self-Supervised Learning},'' \emph{arXiv preprint
  arXiv:2112.03763}, 2021.

\bibitem{nair2021learning}
S.~Nair, E.~Mitchell, K.~Chen, B.~Ichter, S.~Savarese, and C.~Finn,
  ``\href{https://arxiv.org/pdf/2109.01115.pdf}{Learning Language-Conditioned
  Robot Behavior from Offline Data and Crowd-Sourced Annotation},'' in
  \emph{CoRL}, 2021.

\bibitem{shaoconcept2robot}
L.~Shao, T.~Migimatsu, Q.~Zhang, K.~Yang, and J.~Bohg,
  ``\href{http://roboticsproceedings.org/rss16/p082.pdf}{Concept2Robot:
  Learning Manipulation Concepts from Instructions and Human Demonstrations},''
  in \emph{RSS}, 2020.

\bibitem{calvin21}
O.~Mees, L.~Hermann, E.~Rosete-Beas, and W.~Burgard, ``Calvin: A benchmark for
  language-conditioned policy learning for long-horizon robot manipulation
  tasks,'' \emph{IEEE Robotics and Automation Letters (RA-L)}, vol.~7, no.~3,
  pp. 7327--7334, 2022.

\bibitem{andrychowicz2017hindsight}
M.~Andrychowicz, F.~Wolski, A.~Ray, J.~Schneider, R.~Fong, P.~Welinder,
  B.~McGrew, J.~Tobin, P.~Abbeel, and W.~Zaremba,
  ``\href{https://proceedings.neurips.cc/paper/2017/file/453fadbd8a1a3af50a9df4df899537b5-Paper.pdf}{Hindsight
  experience replay},'' in \emph{NeurIPS}, 2017.

\bibitem{tellex2020robots}
S.~Tellex, N.~Gopalan, H.~Kress-Gazit, and C.~Matuszek,
  ``\href{https://h2r.cs.brown.edu/wp-content/uploads/tellex20.pdf}{Robots that
  use language},'' \emph{Annual Review of Control, Robotics, and Autonomous
  Systems}, vol.~3, pp. 25--55, 2020.

\bibitem{lu2019vilbert}
J.~Lu, D.~Batra, D.~Parikh, and S.~Lee,
  ``\href{https://proceedings.neurips.cc/paper/2019/file/c74d97b01eae257e44aa9d5bade97baf-Paper.pdf}{Vilbert:
  Pretraining task-agnostic visiolinguistic representations for
  vision-and-language tasks},'' in \emph{NeurIPS}, 2019.

\bibitem{radford2021learning}
A.~Radford, J.~W. Kim, C.~Hallacy, A.~Ramesh, G.~Goh, S.~Agarwal, G.~Sastry,
  A.~Askell, P.~Mishkin, J.~Clark, \emph{et~al.},
  ``\href{https://arxiv.org/pdf/2103.00020.pdf}{Learning transferable visual
  models from natural language supervision},'' \emph{arXiv preprint
  arXiv:2103.00020}, 2021.

\bibitem{winograd1972understanding}
T.~Winograd, ``Understanding natural language,'' \emph{Cognitive psychology},
  vol.~3, no.~1, pp. 1--191, 1972.

\bibitem{Shridhar-RSS-18}
M.~Shridhar and D.~Hsu,
  ``\href{https://arxiv.org/pdf/1806.03831.pdf}{Interactive Visual Grounding of
  Referring Expressions for Human-Robot Interaction},'' in \emph{RSS}, 2018.

\bibitem{hatori2018interactively}
J.~Hatori, Y.~Kikuchi, S.~Kobayashi, K.~Takahashi, Y.~Tsuboi, Y.~Unno, W.~Ko,
  and J.~Tan, ``\href{https://arxiv.org/pdf/1710.06280.pdf}{Interactively
  picking real-world objects with unconstrained spoken language
  instructions},'' in \emph{ICRA}, 2018.

\bibitem{mees21iser}
O.~Mees and W.~Burgard, ``\href{https://arxiv.org/pdf/2102.08094.pdf}{Composing
  Pick-and-Place Tasks By Grounding Language},'' in \emph{ISER}, 2021.

\bibitem{liu2021structformer}
W.~Liu, C.~Paxton, T.~Hermans, and D.~Fox,
  ``\href{https://arxiv.org/pdf/2110.10189.pdf}{StructFormer: Learning Spatial
  Structure for Language-Guided Semantic Rearrangement of Novel Objects},''
  \emph{arXiv preprint arXiv:2110.10189}, 2021.

\bibitem{shridhar2021cliport}
M.~Shridhar, L.~Manuelli, and D.~Fox,
  ``\href{https://arxiv.org/pdf/2109.12098.pdf}{CLIPort: What and Where
  Pathways for Robotic Manipulation},'' in \emph{CoRL}, 2021.

\bibitem{kingma2013auto}
D.~P. Kingma and M.~Welling,
  ``\href{https://arxiv.org/pdf/1312.6114.pdf}{Auto-encoding variational
  bayes},'' \emph{arXiv preprint arXiv:1312.6114}, 2013.

\bibitem{borja22icra}
J.~Borja-Diaz, O.~Mees, G.~Kalweit, L.~Hermann, J.~Boedecker, and W.~Burgard,
  ``Affordance learning from play for sample-efficient policy learning,'' in
  \emph{ICRA}, 2022.

\bibitem{vaswani2017attention}
A.~Vaswani, N.~Shazeer, N.~Parmar, J.~Uszkoreit, L.~Jones, A.~N. Gomez,
  {\L}.~Kaiser, and I.~Polosukhin, ``Attention is all you need,''
  \emph{NeurIPS}, 2017.

\bibitem{hafner2020mastering}
D.~Hafner, T.~P. Lillicrap, M.~Norouzi, and J.~Ba, ``Mastering atari with
  discrete world models,'' in \emph{ICLR}, 2020.

\bibitem{van2017neural}
A.~Van Den~Oord, O.~Vinyals, \emph{et~al.}, ``Neural discrete representation
  learning,'' \emph{NeurIPS}, 2017.

\bibitem{bengio2013estimating}
Y.~Bengio, N.~L{\'e}onard, and A.~Courville, ``Estimating or propagating
  gradients through stochastic neurons for conditional computation,''
  \emph{arXiv preprint arXiv:1308.3432}, 2013.

\bibitem{salimans2017pixelcnn++}
T.~Salimans, A.~Karpathy, X.~Chen, and D.~P. Kingma, ``Pixelcnn++: Improving
  the pixelcnn with discretized logistic mixture likelihood and other
  modifications,'' \emph{arXiv preprint arXiv:1701.05517}, 2017.

\bibitem{dasari2021transformers}
S.~Dasari and A.~Gupta, ``Transformers for one-shot visual imitation,'' in
  \emph{Conference on Robot Learning}.\hskip 1em plus 0.5em minus 0.4em\relax
  PMLR, 2021, pp. 2071--2084.

\bibitem{bowman2015generating}
S.~R. Bowman, L.~Vilnis, O.~Vinyals, A.~M. Dai, R.~Jozefowicz, and S.~Bengio,
  ``Generating sentences from a continuous space,'' \emph{arXiv preprint
  arXiv:1511.06349}, 2015.

\bibitem{reimers-2019-sentence-bert}
N.~Reimers and I.~Gurevych, ``Sentence-bert: Sentence embeddings using siamese
  bert-networks,'' in \emph{EMNLP}, 2019.

\bibitem{yarats2021drqv2}
D.~Yarats, R.~Fergus, A.~Lazaric, and L.~Pinto, ``Mastering visual continuous
  control: Improved data-augmented reinforcement learning,'' \emph{arXiv
  preprint arXiv:2107.09645}, 2021.

\bibitem{song2020mpnet}
K.~Song, X.~Tan, T.~Qin, J.~Lu, and T.-Y. Liu, ``Mpnet: Masked and permuted
  pre-training for language understanding,'' \emph{NeurIPS}, 2020.

\bibitem{liu2019roberta}
Y.~Liu, M.~Ott, N.~Goyal, J.~Du, M.~Joshi, D.~Chen, O.~Levy, M.~Lewis,
  L.~Zettlemoyer, and V.~Stoyanov, ``Roberta: A robustly optimized bert
  pretraining approach,'' \emph{arXiv preprint arXiv:1907.11692}, 2019.

\bibitem{devlin2018bert}
J.~Devlin, M.-W. Chang, K.~Lee, and K.~Toutanova, ``Bert: Pre-training of deep
  bidirectional transformers for language understanding,'' \emph{arXiv preprint
  arXiv:1810.04805}, 2018.

\bibitem{de2019causal}
P.~de~Haan, D.~Jayaraman, and S.~Levine,
  ``\href{https://proceedings.neurips.cc/paper/2019/file/947018640bf36a2bb609d3557a285329-Paper.pdf}{Causal
  confusion in imitation learning},'' \emph{NeurIPS}, 2019.

\bibitem{mees20icra_asn}
O.~Mees, M.~Merklinger, G.~Kalweit, and W.~Burgard, ``Adversarial skill
  networks: Unsupervised robot skill learning from videos,'' in \emph{ICRA},
  2020.

\end{thebibliography}



\end{document}